\documentclass[10pt,twocolumn,letterpaper]{article}

\usepackage{iccv}
\usepackage{times}
\usepackage{epsfig}
\usepackage{graphicx}
\usepackage{amsmath}
\usepackage{amssymb}
\usepackage{algorithmic}
\usepackage{algorithm}
\usepackage{amsfonts}
\usepackage{appendix}
\usepackage{amsmath}
\usepackage{comment}
\usepackage{float}
\DeclareMathOperator*{\argmax}{argmax} 

\usepackage[pagebackref=true,breaklinks=true,letterpaper=true,colorlinks,bookmarks=false]{hyperref}

\iccvfinalcopy 

\def\iccvPaperID{12439} 

\begin{document}

\title{PrefGen: Preference Guided Image Generation with Relative Attributes}

\author{
Alec Helbling
\and
Christopher J. Rozell\and
Matthew O'Shaughnessy\and
Kion Fallah\and
Georgia Institute of Technology\\
Atlanta, GA 30332
USA
\\
{\tt\small alechelbling@gatech.edu}
}

\maketitle

\begin{abstract}

Deep generative models have the capacity to render high fidelity images of content like human faces. Recently, there has been substantial progress in conditionally generating images with specific quantitative attributes, like the emotion conveyed by one's face. These methods typically require a user to explicitly quantify the desired intensity of a visual attribute. A limitation of this method is that many attributes, like how ``angry'' a human face looks, are difficult for a user to precisely quantify. However, a user would be able to reliably say which of two faces seems ``angrier''. Following this premise, we develop the \textit{PrefGen} system, which allows users to control the relative attributes of generated images by presenting them with simple paired comparison queries of the form \textit{``do you prefer image $a$ or image $b$?''} Using information from a sequence of query responses, we can estimate user preferences over a set of image attributes and perform preference-guided image editing and generation. Furthermore, to make preference localization feasible and efficient, we apply an active query selection strategy. We demonstrate the success of this approach using a StyleGAN2 generator on the task of human face editing. Additionally, we demonstrate how our approach can be combined with CLIP, allowing a user to edit the relative intensity of attributes specified by text prompts. \footnote{Code at https://github.com/helblazer811/PrefGen}

\end{abstract}

\section{Introduction}
\label{sec:intro}

Deep generative models have demonstrated the ability to generate high-quality images of content like human faces \cite{deep_gen_survey}. To expand their capabilities, there is substantial interest in improving the \textit{controllability} of these models, providing the end user the ability to modify the continuous attributes of output images \cite{ganspace, shen2020interpreting, Abdal_2021, Nie2021ControllableAC, shoshan2021gan}. Some of these techniques allow a user to edit a set of visual attributes found through unsupervised learning \cite{ganspace, shen2020interpreting}, while others leverage supervised learning or pre-trained classifiers to form mappings from attributes to the generative model latent space \cite{Abdal_2021, Nie2021ControllableAC, shoshan2021gan}. However, with all of these approaches, if a user wants to precisely modify the intensity of a visual attribute, they must explicitly tune a set of quantitative attributes. This user interface is ill-suited to the task of modifying the relative attributes \cite{relative_attributes} of images, like how ``sad'' or ``tired'' a person looks. Unlike properties such as ``temperature,'' people don't share a common notion of scale for attributes like ``anger'' \cite{Thurstone}, making it difficult for them to explicitly quantify how ``angry'' a face looks. This limits the utility of attribute-conditioned generative models to the class of attributes that can be easily quantified. 

\begin{figure}
    \centering
    \includegraphics[width=\columnwidth]{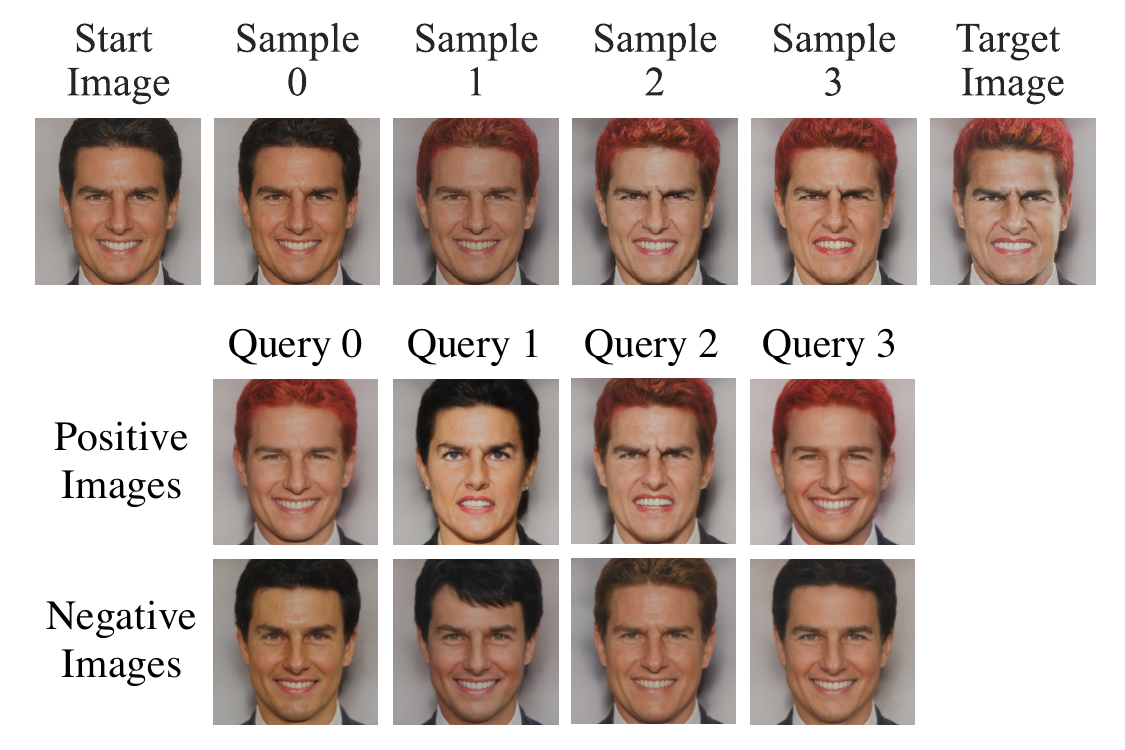}
    \caption{\textbf{Preference estimation of relative textual attributes.} We estimate an oracle's preferences over the textual attributes ``anger" and ``red hair" using \textit{PrefGen}. Given a start image, PrefGen poses a sequence of pairwise comparisons to determine an oracle's preferred attributes in a target image. The top row shows the generated estimate of an oracle's preference after answering each query. The second and third row represent the query prompted to the oracle, with the images in the second row being preferred to those in the third. As more queries are answered, the image generated from estimated preference converges to the ground truth target in the final column.}
    \label{fig:intro_figure}
\end{figure}

\begin{figure*}
    \centering
    \includegraphics[width=6.5in]{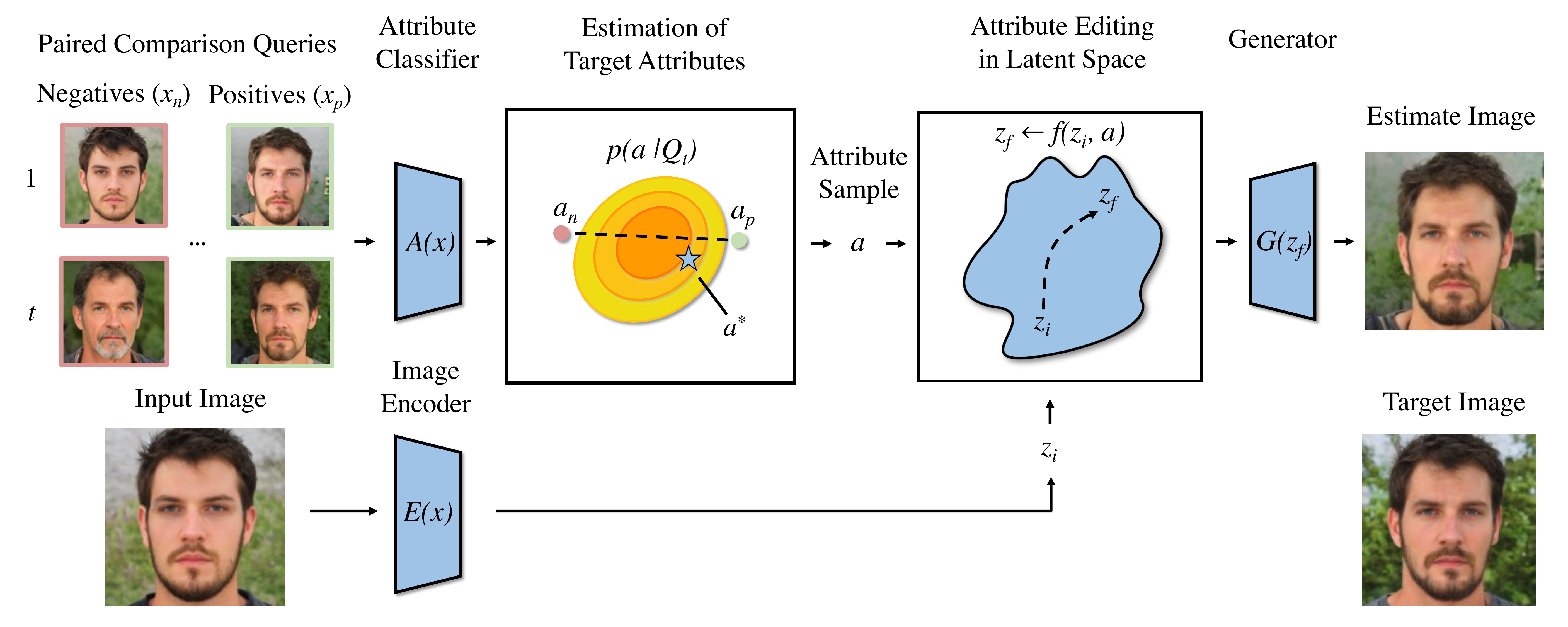}
    \caption{\textbf{Proposed system for preference guided image editing.} After mapping an image to the attribute space with $A(x)$, we present an oracle with a sequence of paired comparison queries $Q_t = \{q_1 \dots q_t \}$. Responses to the paired queries are used to compute the posterior distribution over the attribute space $p(a \mid Q_t)$. We sample from this posterior distribution to produce an  estimate of the user's ideal point $a^*$. We combine the ideal point estimate with the latent vector of the input image $E(x_i) = z_i$ to produce a final vector in the generative model's latent space $z_f \gets f(z_i, a)$. Finally, we map our modified latent vector $z_f$ to image space $G(z_f)$ using our generator. }
    \label{fig:explanatory}
\end{figure*}

While people may not be able to explicitly specify how ``angry'' a face looks, they can more easily say which of two faces looks ``angrier'' \cite{kendall1940method}. Motivated by this observation, we develop the \textit{PrefGen} system for controlling the relative attributes \cite{relative_attributes} of images using the information from paired comparison queries of the form \textit{``do you prefer image $x_a$ or image $x_b$?''} By presenting a user with a sequence of these queries, we can deduce their preferences over a set of relative attributes without the need to explicitly quantify them.
We leverage the information from a sequence of these queries $Q$ to infer a distribution of user preferences over a space of relative attributes $\mathcal{A}$. We use a Bayesian framework for the attribute preferred by the user, allowing us to actively select queries \cite{settles2012active, canal.18d, jamieson2011active} and quantify uncertainty. This allows us to efficiently search the space of attributes as well as account for inconsistencies in a user's responses, respectively.

Paired comparisons have long been regarded as a rich mode of information for inferring user preferences over a set of items \cite{kendall1940method} as well as reasoning about difficult-to-quantify psychological scales \cite{Thurstone}. Additionally, paired comparison queries can be passively extracted from widely available sources like search engine mouse click data \cite{joachims_clickthrough}. This provides the potential for the proposed framework to control a generative model's output using implicit feedback from users, rather than requiring them to explicitly quantify their preferences \cite{ganspace, shen2020interpreting, Abdal_2021, Nie2021ControllableAC, shoshan2021gan}.

In our experiments, we demonstrate the success of our method with the StyleGAN2 \cite{karras2019style} architecture, a widely used Generative Adversarial Network (GAN) \cite{goodfellow2014generative} architecture. However, in principle, our method is sufficiently general to be connected with any generative model with the ability to conditionally generate images with different quantitative attributes. We apply our methodology to the task of editing the attributes of human faces. We show that PrefGen can be easily used in conjunction with existing approaches for attribute-conditioned image generation. Specifically, we train supervised mappings from attributes, like face age and pose, to latent vectors, as is done in \cite{shoshan2021gan}, and infer user preferences over those attributes using paired comparisons. Recent works show that it is possible to modify the visual attributes of images from textual information \cite{styleclip}. Building on this work, we show that you can leverage joint vision-language models like CLIP \cite{radford2021learning} to edit the relative intensity of visual attributes using the information from paired comparisons.

The contributions of this work are:
\begin{itemize}
    \item The PrefGen system, which adds controllability to a pretrained generative model by allowing users to edit the relative attributes of an output image by answering a sequence of pairwise queries.
    \item Empirical validation with two methods for building an attribute encoder, with an extension to previous methods for active query selection that leverages a continuous generative attribute space.
\end{itemize}

\section{Background and Related Works}

\paragraph{Relative Attributes}

Many attributes have an inherent ordering, but are difficult for people to explicitly quantify. For example, people can sort clothing based on how ``fashionable'' they look and have an understanding of how ``fancy'' a dinner is, but would likely struggle to consistently quantify these attributes. These are called relative attributes \cite{relative_attributes} and there is a substantial body of work attempting to model them. Much work focuses on learning to rank the relative intensity of visual attributes \cite{liu2009learning, yang2016deep, singh2016end}. A related line of work focuses on learning metrics of similarity or \textit{similarity embeddings}, where distances between items in the embedding space correspond to notions of similarity \cite{tamuz2011adaptively, agarwal2007generalized, facenet}. The authors of WhittleSearch \cite{whittlesearch} have a similar motivation to us, where they wish to search a finite set of images using relative attribute feedback (paired comparison queries). However, our approach allows for sampling from a continuous space of attributes using a generative model, and not just a finite set of images.  

\begin{figure}[t!]
    \includegraphics[width=\columnwidth]{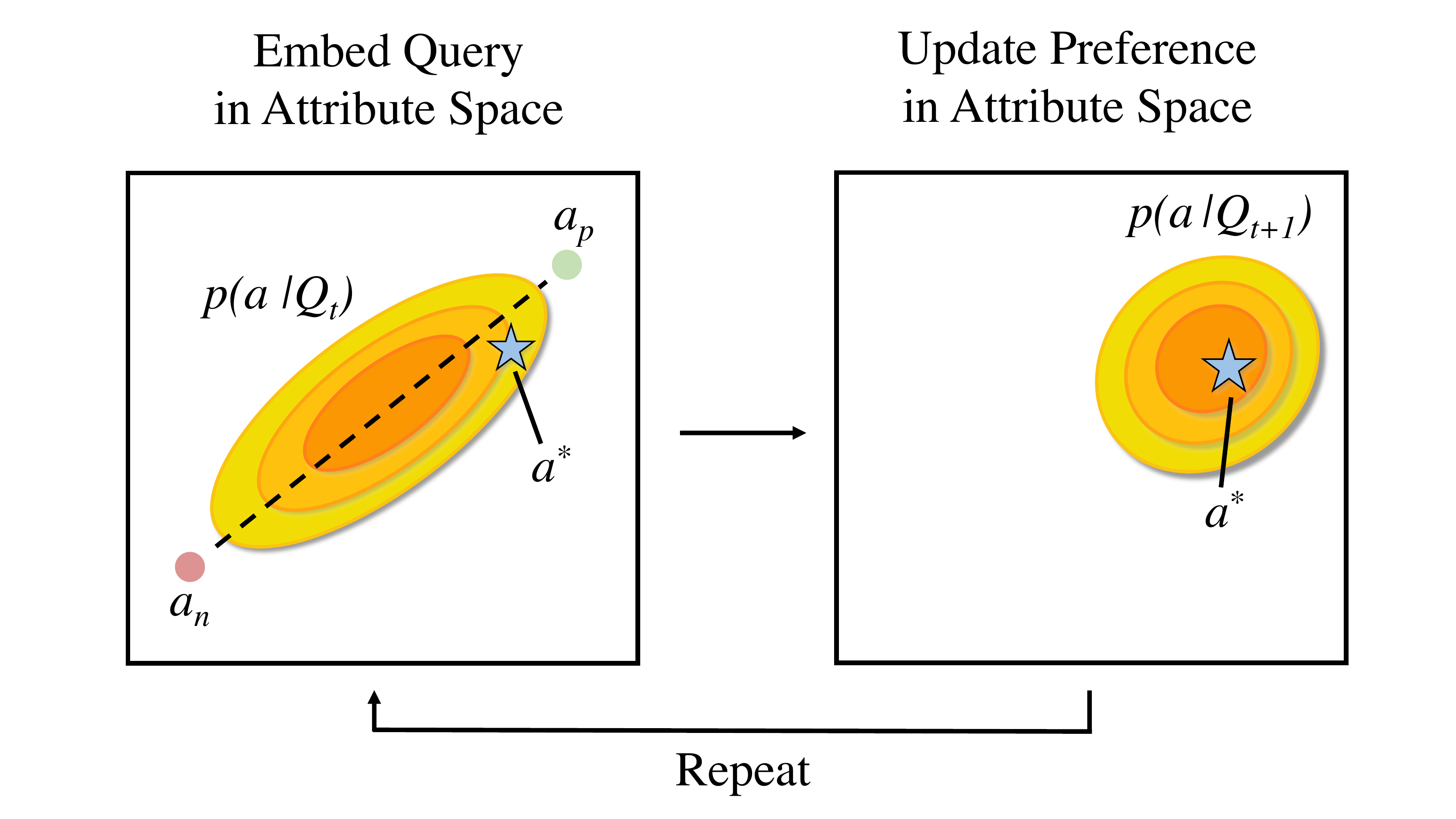}
    \caption{\textbf{Preference estimate update after a paired query.} Presenting a user with a paired query $q$ allows us to estimate a distribution over user preference $p(a \mid q)$. After asking a query $q$, we can combine that information with the set of previous queries $Q_t$ such that $Q_{t+1} = Q_{t} \cup q$ and narrow down a region in attribute space that is likely to align with user preference.}
    \label{fig:localization}
\end{figure}

\paragraph{Controllable Generative Models}

A difficult and open-ended problem in generative modeling is how to control the attributes of generated content. Many generative models have the capacity to generate high fidelity examples \cite{goodfellow2014generative, latent_diffusion}, but require extra methodology to control. There is a significant body of work focused on conditional sampling of images with various continuous attributes using GANs \cite{Nie2021ControllableAC, Abdal_2021, shen2020interpreting, styleclip},  without modifying the weights of the underlying GAN. Some approaches deploy supervised mappings with various underlying architectures, like normalizing flows \cite{Abdal_2021} and energy-based models \cite{Nie2021ControllableAC}. Other methods attempt to infer interpretable directions in StyleGAN's latent space \cite{shen2020interpreting}. \cite{shoshan2021gan} train neural network mappings from continuous attributes to StyleGAN latent space, and fine-tune its representation so that it is disentangled (i.e. different latent subspaces correspond to different interpretable attributes). All of these methods require users to explicitly quantify continuous attributes. However, many attributes are relative, and are difficult for user's to explicitly quantify. Motivated by this problem, we develop a method for user's to control the attributes of generated images by responding to paired comparison queries. In our experiments, we demonstrate how our approach can be easily deployed on top of existing conditional generation methods like \cite{shoshan2021gan} to conditionally generate images with different continuous attributes using the information from paired queries. 

\paragraph{Text-driven Image Generation}

There have been substantial recent advancements in guiding the outputs of generative models using text \cite{glide, ramesh2021zeroshot}. In the StyleGAN literature, there is a body of work focusing on finding directions in the StyleGAN latent space that correspond to different textual attributes \cite{styleclip, Abdal_2021}. While supervised methods for conditional generation of images with different attributes have impressive capabilities, they require the use of large sets of annotated data or pre-trained classifiers. This limits the user's editing capabilities to a narrow set of pre-defined attributes. In StyleCLIP \cite{styleclip}, the authors show that it is possible to edit StyleGAN generated images to match certain text prompts like ``an angry person'' or ``a cute cat.'' They use the CLIP \cite{radford2021learning} joint text-image embedding architecture to infer directions in StyleGAN latent space that correspond to different textual attributes. Language provides a fixed lexicon of adjectives and modifiers for describing the intensity of visual attributes. This can make it challenging to precisely control the relative intensity of an attribute with text alone. In our experiments, we deduce directions in CLIP space using an approach similar to StyleCLIP, and demonstrate that it is possible to use \textit{PrefGen} to modify the intensity of relative textual attributes.

\paragraph{Estimating Preferences with Paired Comparisons}

Paired comparison queries of the form \textit{``do you prefer image $x_a$ or image $x_b$''} offer an elegant way to model a user's preferences over relative attributes \cite{furnkranz2010preference, aggarwal2019modelling}. They unlock the ability to model intuitive concepts that are difficult to explicitly describe or quantify \cite{david1963method, Thurstone}. Additionally, paired comparison queries can be passively collected from widely available data, like search engine clicks \cite{joachims_clickthrough, knowing_every_move}. This means that it is possible to synthesize user preference information from abundantly available sources and use generative models to generate content that matches user preferences. 

Our work closely draws on methods for efficiently searching pre-existing similarity embeddings using the information from paired comparisons \cite{canal.18d, jamieson2011active, davenport2013lost, xu2020simultaneous}. A common simplifying assumption in this literature is to represent the set of features or attributes that a user prefers by a single point, called the \textit{ideal point} $a^* \in \mathcal{A}$ \cite{Coombs1950-COOPSW}. We assume that given two points in an embedding, a user would prefer the one closer to $a^*$. Our goal is to estimate this ideal point using the information from paired queries. The number of possible paired queries scales on the order of $O(N^2)$ with the number of images $N$, so it is typically infeasible and redundant to present a user with all possible queries. The authors of \cite{jamieson2011active} devise an approach to more efficiently rank a fixed set of items by actively selecting paired queries. Other approaches build on this methodology by attempting to search for the ideal point over a continuous embedding \cite{davenport2013lost, xu2020simultaneous, canal.18d}. 

\cite{GaussianPreferenceLearning} devise a probabilistic approach for estimating preferences and develop methods for actively selecting queries based on information theory. We most closely draw from the work of \cite{canal.18d}, which devises metrics for actively selecting queries from a fixed set in a way that accounts for noise in their responses. We apply their approach to the task of estimating user preferences over a space of relative visual attributes, and build upon it by selecting optimal queries from our continuous space instead of from a fixed set. To the best of our knowledge, our work is the first that utilizes paired comparison search with active querying to control the attributes of a generative model. 

Perhaps most similar to our approach is the work of \cite{heim2019constrained}, where the authors attempt to control a GAN given a set of paired comparison constraints. However, they leverage a long short-term memory (LSTM) network to directly predict a latent vector that generates an image that satisfies pairwise constraints. We empirically show that this approach leads to inferior performance over an analytic Bayesian framework \cite{GaussianPreferenceLearning, canal.18d}, potentially due to the ill-posed problem of generalizing from training queries to a specific user's query responses. We adapt this approach as a baseline in Section~\ref{sec:exp}. 

\section{Methods}

\begin{figure}
    \centering
    \includegraphics[width=\columnwidth]{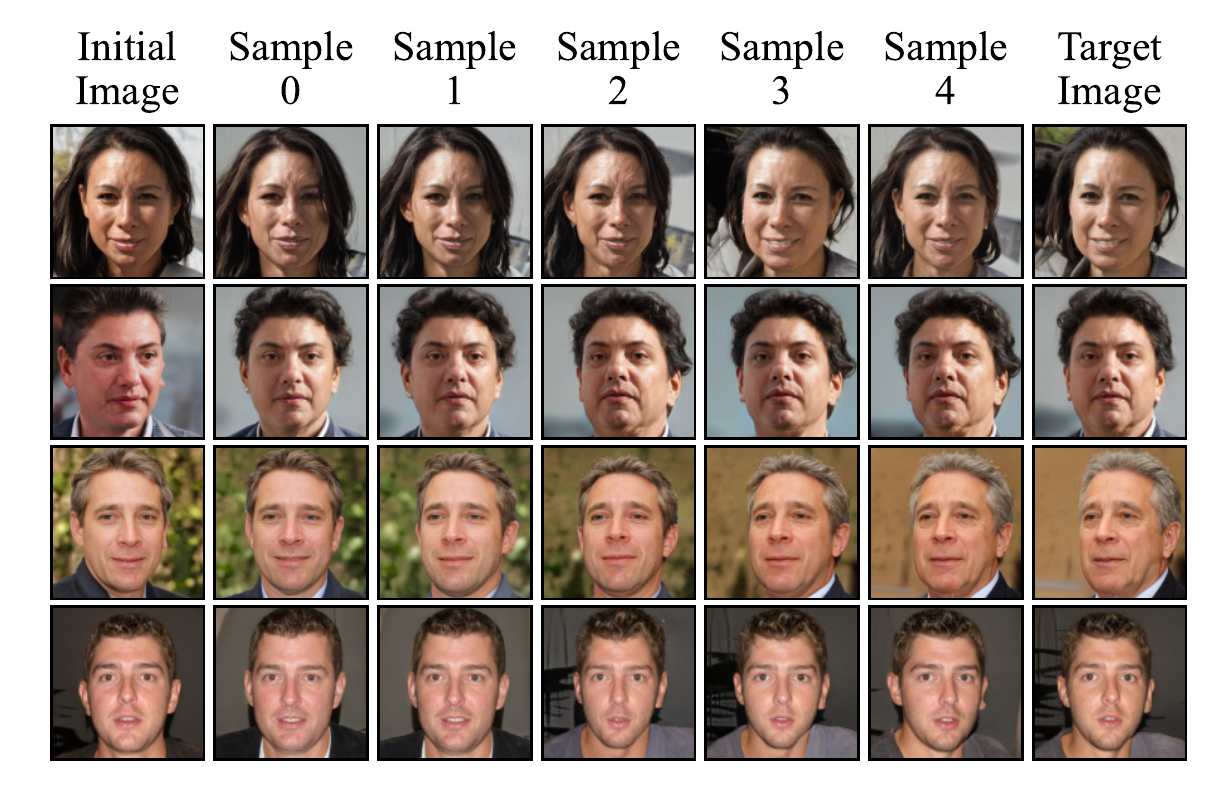}
    \caption{\textbf{Edited image over queries collected.} Images generated to match oracle preferences after N queries. Over time, the generated samples change from the input image to more closely resemble the target image in the last column. }
    \label{fig:preference_estimation_just_samples}
\end{figure}

\begin{algorithm}[t]
    \caption{Paired Comparison Based Image Generation}
    \label{alg:localization}
    \begin{algorithmic}
        \STATE {\bfseries Input:} Input Image $x_{i}$
        \STATE Encode the input image to latent space $z_{i} = E(x_{i})$
        \STATE Initialize query set $Q_0 = \{\}$
        \FOR{$t=1$ {\bfseries to} $T$}
        \STATE Actively select a query $q = S(Q_t)$
        \STATE Prompt the oracle for response to $q$
        \STATE Add the answered query to $Q_t = Q_{t-1} \cup q$
        \ENDFOR
        \STATE Compute $\widehat{a}^* = \mathbb{E}[a \mid Q_T]$ using samples from MCMC.
        \STATE Modify $z_{i}$ with our estimate $z_{f} = f(z_{i}, \widehat{a}^*)$
        \STATE Generate the final edited image $x_{f} = G(z_{f})$
        \STATE {\bfseries Output:} $x_{f}$
        \end{algorithmic}
\end{algorithm}

Our goal is to allow an oracle to control the relative attributes of generated images by asking simple paired comparison queries of the form \textit{``do you prefer image $x_a$ or image $x_b$?''} Using the information from a sequence of these queries $Q_t = \{q_1, \dots, q_t\}$ we want to estimate an oracle's preferences over a set of relative attributes $\mathcal{A}$. We can present an unanswered query $q_i = (x_a, x_b)$ to an oracle $O(q_i)$ which returns an answered query $O(q) = (x_p, x_n)$. Thus, 
\begin{equation}
    q_i = (x_p, x_n) \iff x_p \prec x_n,
\end{equation}
where $x_p \prec x_n$ indicates that oracle ``prefers'' the image $x_p$ to the image $x_n$. We assume that there exists some vector $a^* \in \mathcal{A}$ called the \textit{target attribute} which best represents an oracle's preferences. Our goal is to infer $a^*$ from the information in queries $Q_t$.

Once we have an estimate $\widehat{a}^*$ of the oracle's ideal point, we need a way to generate images that are consistent with those preferences. We do this by mapping the attribute to the generative model latent space $\mathcal{Z}$. We represent this with an \textit{attribute mapping} function, from a latent vector $z_i$ and attribute $a$ to another latent vector $z_f \gets f(z_i, a)$. We can both modify an initial latent vector $z_i$, perhaps found using GAN Inversion $z_{i} = E(x_{i})$ \cite{encoder4editing}, or generate images using only the information from our attribute $z_i \sim p(z)$. Our goal is to model $f(z_i, a)$ so that $A(G(z_{f})) \approx a$ and the identities of $G(z_{f})$ and $G(z_{i})$ are the same, where $A(x)$ is a function that predicts the attributes of an image $x \in \mathcal{X}$. We explore two key ways of modeling $A(\cdot)$ and $f(\cdot)$ in Section \ref{controlling_attributes}.

\begin{figure*}[th!]
    \centering
    \includegraphics[width=3.25in]{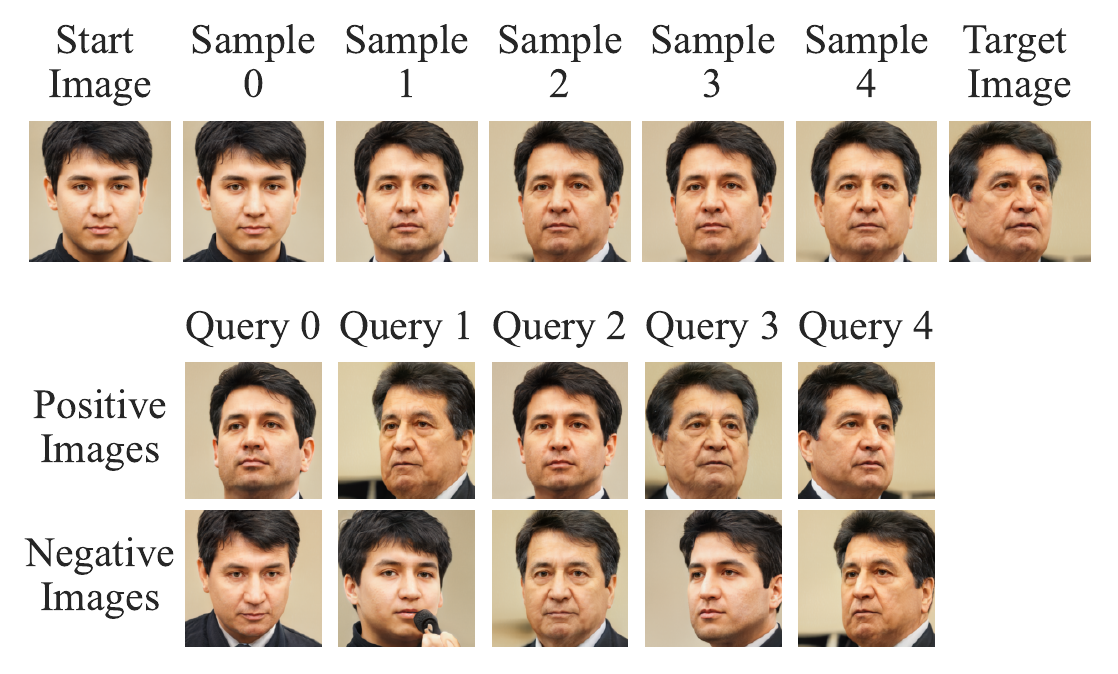}
    \hspace{.3in}
    \includegraphics[width=3.25in]{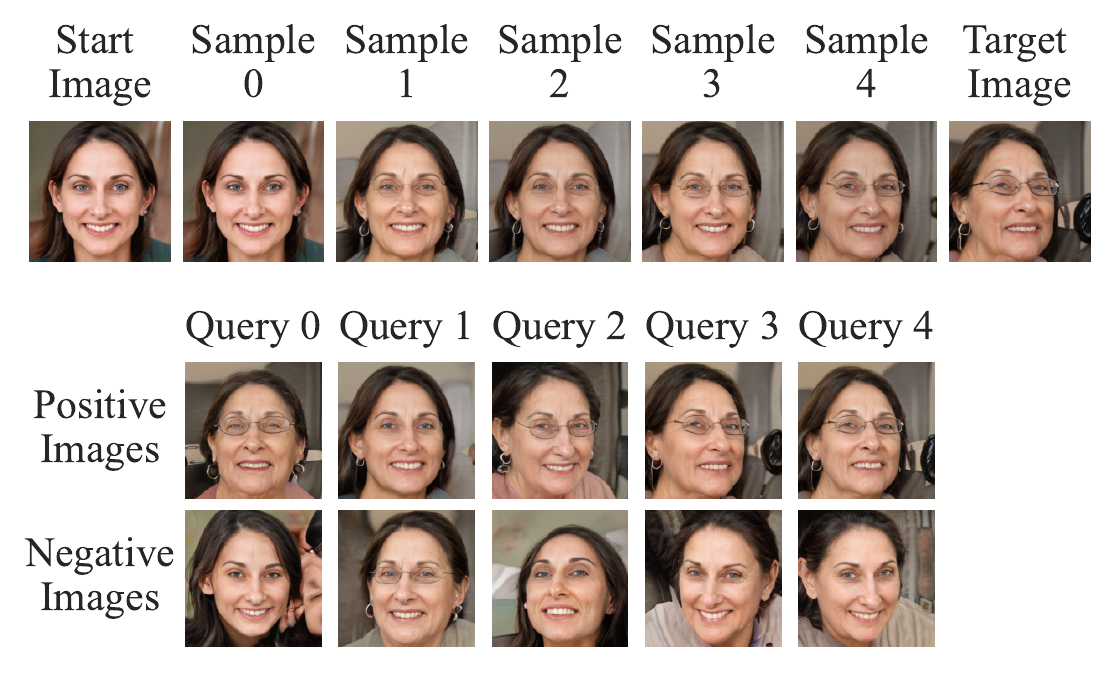}
    \caption{\textbf{Qualitative trial run of PrefGen.} We demonstrate two trial runs of PrefGen in a 4D space formed by yaw, pitch, and age of human faces using a supervised attribute mapping. See Figure~\ref{fig:intro_figure} for figure details. We show each of the queries presented to the oracle, and images generated by sampling from the posterior after each query. Sample images qualitatively converge to the target image as query responses are collected.}
    \label{fig:qualitative_gan_control_with_queries}
\end{figure*}

\subsection{A Bayesian Model of User Preferences}

We use a Bayesian framework to estimate the distribution $p(a \mid Q_t)$ of user preferences over attribute space $\mathcal{A}$ using the information from the queries $Q_t$. We use a simplifying assumption, called the ideal point model \cite{Coombs1950-COOPSW}, that a user's preference for a set of attributes $a \in \mathcal{A}$ corresponds to the distance between that vector $a$ and the target attributes $a^*$. Let $A(x)$ be a (perhaps unknown) mapping from image space $\mathcal{X}$ to attributes $\mathcal{A}$. We say a user ``prefers'' a positive image $x_p$ to a negative image $x_n$ when the attributes $a_p = A(x_p)$ of the positive image $x_p$ are closer to the target attributes $a^*$ than the attributes $a_n = A(x_n)$ of the negative image $x_n$:
\begin{equation}
    x_p \prec x_n \iff \Vert a^* - a_p \Vert < \Vert a^* - a_n \Vert.
    \label{eq:similarity_embedding_assumption}
\end{equation}

Following the assumption in Equation \ref{eq:similarity_embedding_assumption} and work from \cite{canal.18d} we represent the likelihood that $x_p \prec x_n$ as
\begin{equation}
    p(q \mid a) = \sigma\bigl(k_q (\Vert a -a_n \Vert ^2 - \Vert a - a_p \Vert ^2)\bigr)\label{logistic_likelihood},
\end{equation}
where $\sigma$ is the logistic function and $k_q$ is a scalar value that accounts for the noisiness of oracle responses. Intuitively, this likelihood produces a very high probability when a user prefers an image with attributes $a_p$ that is much closer to the ideal point $a^*$ than the image with attributes $a_n$. The noise constant $k_q$ can be thought of as a signal-to-noise ratio that determines the expected amount of information contained in each query. It is possible to for the oracle to accommodate for an expectation of different amounts of noise in different queries by changing $k_q$ with different queries \cite{canal.18d}. 
 
By representing this problem probabilistically, we can accommodate various sources of error, such as ``noise'' in user responses to paired queries. We can combine the information from multiple paired queries through a recursive application of Bayes rule,
\begin{equation}
    p(a \mid q_1, ..., q_t) = \frac{p(a) \prod_{j=1}^{t}p(q_j \mid a)}{p(q_t \mid q_1, ..., q_{t-1})},
\end{equation}
where our prior over the user preference $p(a)$ is a uniform distribution. Finally, we use MCMC to sample from the posterior distribution $p(a \mid Q_t)$ where $Q_t = \{q_1, \dots, q_t\}$ and take the expected value $\mathbb{E}[a \mid Q_t]$ to form an estimate of the target attributes $a^*$. Intuitively, this process can be thought of as a multidimensional generalization of the binary search algorithm. This procedure is shown in Algorithm \ref{alg:localization} and Figure \ref{fig:localization}.

\begin{figure*}
    \centering
    \includegraphics[width=\columnwidth]{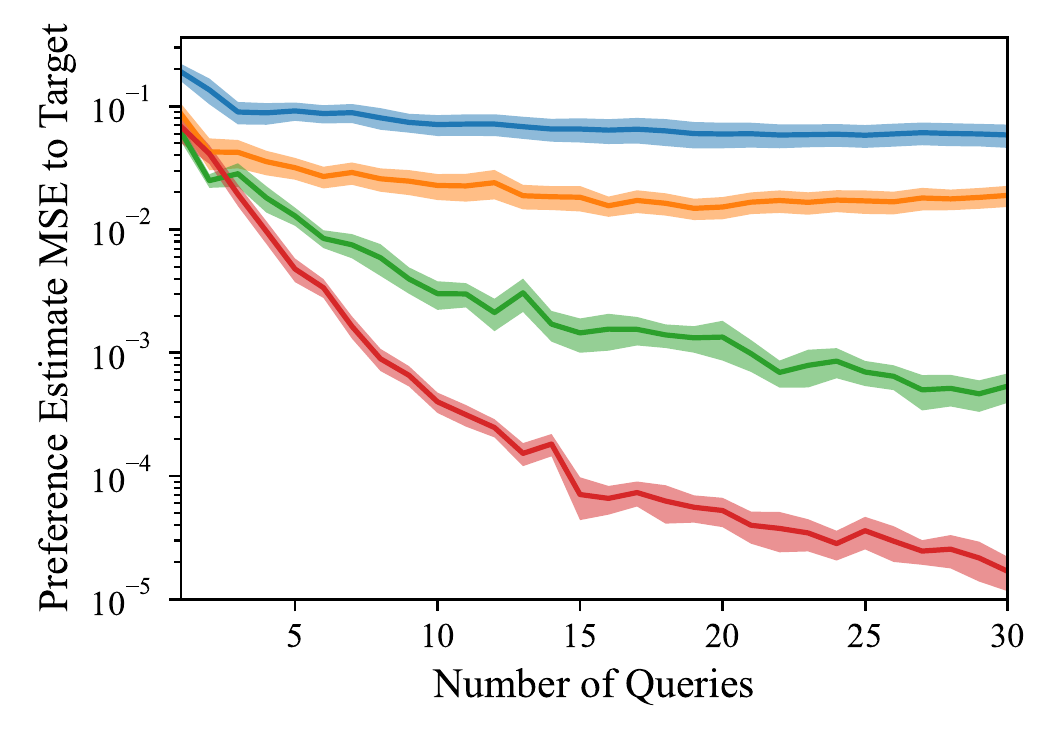}
    \includegraphics[width=\columnwidth]{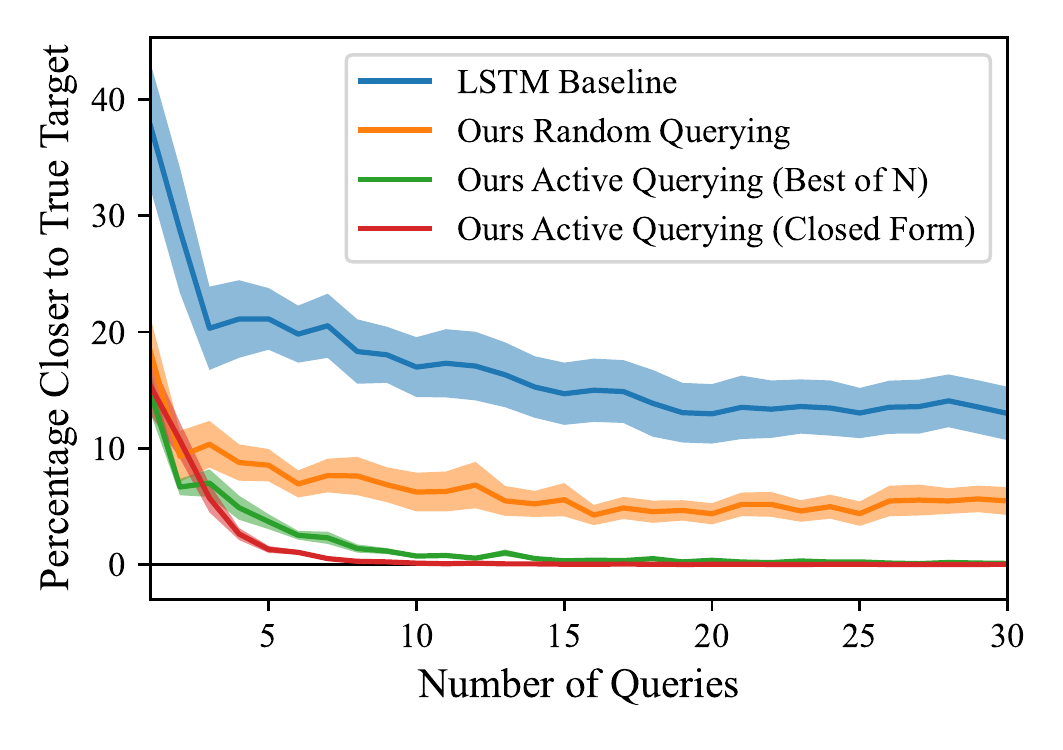}
    \caption{\textbf{Comparing the quantitative performance of various approaches to estimating user preferences from paired comparisons}. We simulate an oracle's response to a sequence of queries for 20 trials, with 30 queries per trial. We search a 2D space formed by the age and the yaw (left-right pose) of a face. (Left) MSE in attribute space of preference estimate after $t$ queries and the ground truth attribute $a^*$ for a given trial. (Right) Percentage of attribute vectors from a batch of 1000 that are closer to the target attributes than the estimate. According to both metrics, PrefGen is able to efficiently estimate preferences with high precision. As expected, active querying improves the efficiency of preference estimation as compared to randomly querying, and the \textit{Closed Form} approach, which leverages the continuous generative model latent space, significantly outperforms the \textit{Best of N} approach.}
    
    \label{fig:comparison_figure}
\end{figure*}

\subsection{Actively Query Selection for Efficient Preference Estimation \label{active_selection}}

It is generally infeasible to present a user with every possible paired comparison query. Given a fixed set of $N$ images, one can construct $O(N^2)$ paired queries from them. Furthermore, in our setting we have a generative model capable of continuously sampling from attribute space, so we have an infinite number of possible queries to choose from.  So the problem stands, how do we select a set of queries to best estimate the user's preferences? A naive solution, which we call \textit{random querying}, is to simply select a small, random subset of the many possible queries. However, many queries are redundant given the information from previous queries that were presented to a user. 

We build on methods from active learning \cite{settles2012active} to select queries that minimize the amount of uncertainty in our posterior distribution $p(a \mid Q_t)$. The authors of \cite{canal.18d} develop strategies for selecting queries that contain as much new information as possible about the ideal point $a^*$ from a discrete set of candidate queries. One such strategy is called mean-cut max-variance (MCMV). Intuitively, the goal of MCMV is to select a query $q = (a_p, a_n)$ along an axis of high variance as determined by the posterior $p(a \mid Q_t)$, and that bisects the mean of the posterior $\mathbb{E}[a \mid Q_t]$. We can represent the plane that bisects the attributes $a_p$ and $a_n$ with a normal vector $v = 2(a_p - a_n)$ and scalar offset $\tau = |a_p|^2 - |a_n|^2$. We can compute the variance of the posterior distribution in the direction of the query computed by $\sigma_q = v \Sigma_{a} v^T$ where $\Sigma_{a}$ is the covariance of samples from our posterior. Cutting the mean means selecting a query $q = (a_p, a_n)$ such that the plane between the two query items parameterized by $v$ and $\tau$ intersects with the mean of the distribution. This can be computed as $\mu_q = (v^T \mathbb{E}[a \mid Q_t] - \tau) / |v|$. So the best query is the one that minimizes $k_q \sigma_q - \lambda \mu_q$, where $k_q$ is our aforementioned noise constant and $\lambda$, is a hyperparameter used to balance the relative importance of the mean-cut and maximum-variance criteria. 

In practice, we need to sample many potential queries from our attribute space, measure the MCMV utility of each, and select the query with the highest value. As we converge on an estimate of the oracle's preferences, performance starts to degrade since the best query is farther from the mean and axis of maximum variance relative to the volume of the posterior (See Figure \ref{fig:comparison_figure}). Our generative modeling framework grants us the ability to sample continuously from attribute space (See Section \ref{controlling_attributes}). This enables us to directly select a query that is mean-cut and max-variance, obviating the need to sample large numbers of candidate queries. We simply select a query with points equidistant to the mean that are along the axis of maximum variance of our posterior $p(a \mid Q_t)$. It is worth noting that this approach does not account for models of noise that may interfere with the expected informativeness of different queries. We call this approach \textit{active closed-form} and the original MCMV approach \textit{active best of N}.

\subsection{Controlling the Attributes of Generated Images \label{controlling_attributes}}

In this paper, we explore two key approaches for modeling the attribute mapping function $f(z_i, a)$. It is important to note that these two approaches are just two of many possible ways one could construct a mapping from attribute vectors to latent vectors (or directly images).

\paragraph{Supervised Attribute Mapping: Constructing Supervised Mappings from Attributes to Latent Space}

For our first method we train supervised mappings from attributes like face pose and age to our generator's latent space. For the implementation of this approach, we build on top of GANControl \cite{shoshan2021gan}, where the authors train supervised mappings from attributes to a disentangled StyleGAN latent space. This work shows that it is possible to modify quantitative attributes without significantly changing the underlying identity of images. 

\paragraph{CLIP Attribute Mapping: Modifying Relative Textual Attributes with StyleCLIP}

In StyleCLIP \cite{styleclip}, the authors introduce a method to find a latent vector that generates an image with a certain attribute, without altering the identity of a given input latent vector. They describe a gradient based approach for finding latent vectors that generate images with certain textual attributes. Building off of StyleCLIP, we encode model relative attributes, allowing us to interpolate between two text prompts: a neutral text prompt like \textit{``a person with a neutral expression''} and a target text prompt ``\textit{a person with an angry expression''}. We can embed these prompts into CLIP \cite{radford2021learning} space using the CLIP text encoder $\mathcal{T}(t)$ to produce a neutral and target vector $t_n$ and $t_t$ respectively.  Images with cosine similarity close to $t_n$ will have neutral expressions, and those closer to $t_t$ will better match the target prompt. Let  $\mathcal{V}(x)$ be the CLIP vector of an image $x \in \mathcal{X}$. We can predict the intensity of an attribute by computing the cosine similarity between $\mathcal{V}(x)$ and our attribute direction $t_a = t_t - t_n$ (i.e. $\frac{\langle \mathcal{V}(x), t_a \rangle}{\Vert\mathcal{V}(x)\Vert\Vert t_a \Vert}$). We can then modify an image's attribute by simply choosing some $\alpha \in [0, 1]$ and minimizing the distance between our current predicted attribute intensity and $\alpha$. We do this using gradient descent as is done in StyleCLIP with additional penalty terms to ensure the identity of the image stays the same. This approach allows us to quantitatively model the intensity of a relative attribute, which is difficult to do by directly encoding text prompts as is done in \cite{ramesh2021zeroshot}.
\section{Experiments}
\label{sec:exp}

The focus of our experiments is to evaluate the ability of our proposed method to effectively estimate user preferences over a space of attributes from paired comparison queries and conditionally generate images that match those preferences. A core evaluation method that we use is to simulate user responses to queries with a simple oracle. For each preference estimation trial, we sample some arbitrary target attribute vector $a^* \in \mathcal{A}$ uniformly from $[0, 1]^d$, which corresponds to the $d$ image attributes that our user wants to be in samples images. We then sequentially pose several paired comparison queries, each of the form $q = (a_p, a_n)$, such that they satisfy Equation \ref{eq:similarity_embedding_assumption}. 

\subsection{PrefGen with a Supervised Attribute Mapping}

\paragraph{Implementation Details} 
One way to represent $f$ is to train a neural network with supervised learning that maps attributes to the latent space of a generative model. We use the methods from GANControl \cite{shoshan2021gan} to construct supervised mappings from the pose (yaw, pitch, and roll) and the apparent age of a face to the latent space of a StyleGAN2 \cite{stylegan_original} using the FFHQ Human Attributes dataset \cite{karras2019style}. In our quantitative experiments, for each method we conduct 20 trials of preference estimation, each with 30 queries and $k_q = 10$.

\paragraph{Quantitative Evaluation of PrefGen}

\begin{figure}[t]
    \centering
    \includegraphics[width=\columnwidth]{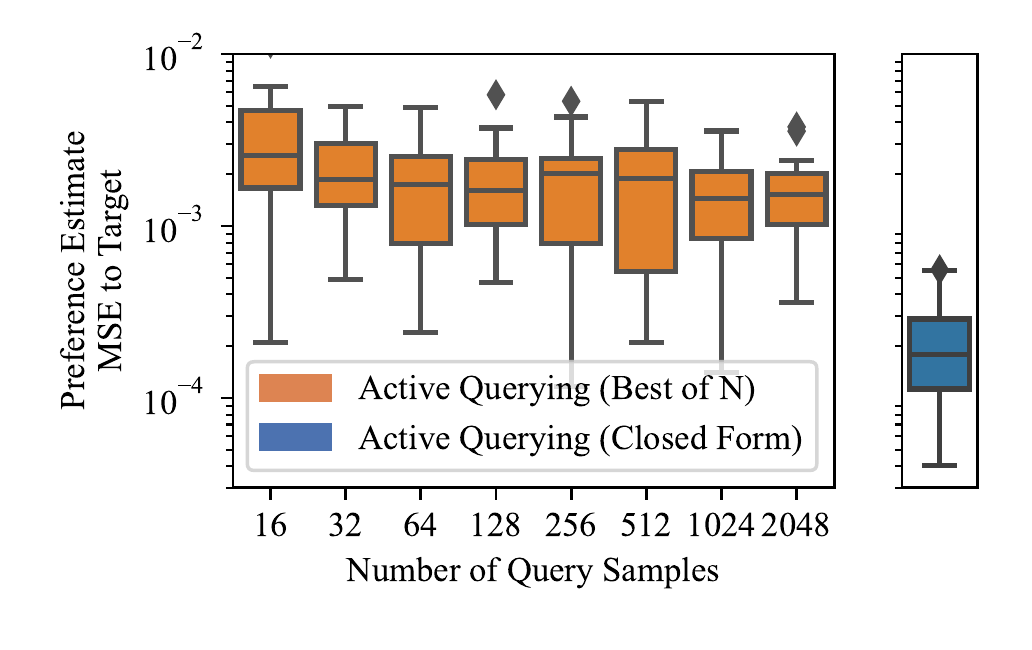}
    \caption{\textbf{Improving active selection.} (Left) Preference estimation performance with 30 query responses as the number of candidate queries increases using \textit{Best of N}. (Right) Even as the number of candidate queries increases, the \textit{Best of N} strategy has higher attribute estimation error compared to the proposed \textit{Closed Form} strategy. }
    \label{fig:closed_form_comparison}
    \vspace{-.1cm}
\end{figure}

To measure our ability to estimate a user's target attributes $a^*$ both precisely and efficiently, we use three key metrics: (1) the Mean Squared Error (MSE) of our current estimate and the ground truth attribute in the attribute space after each query response, (2) the percentage of attributes closer than our estimate to the target attributes $a^*$, and (3) the percentage of paired comparison constraints where the estimate of the target attributes $a^*$ is closer to the positive than the negative. Figure \ref{fig:comparison_figure} shows that our approaches (both active and random) are able to converge on preference estimates with low MSE to the target attributes and that are closer than a large percentage of randomly sampled attribute vectors. Table \ref{comparison_table} further confirms these results, and shows that \textit{PrefGen} can estimate a set of attributes that satisfy a high percentage of constriants. Further results in Appendix \ref{further_results} show that this pattern is also replicated when searching over higher dimensional attribute spaces. 

\begin{figure}
    \centering
    \includegraphics[width=\columnwidth]{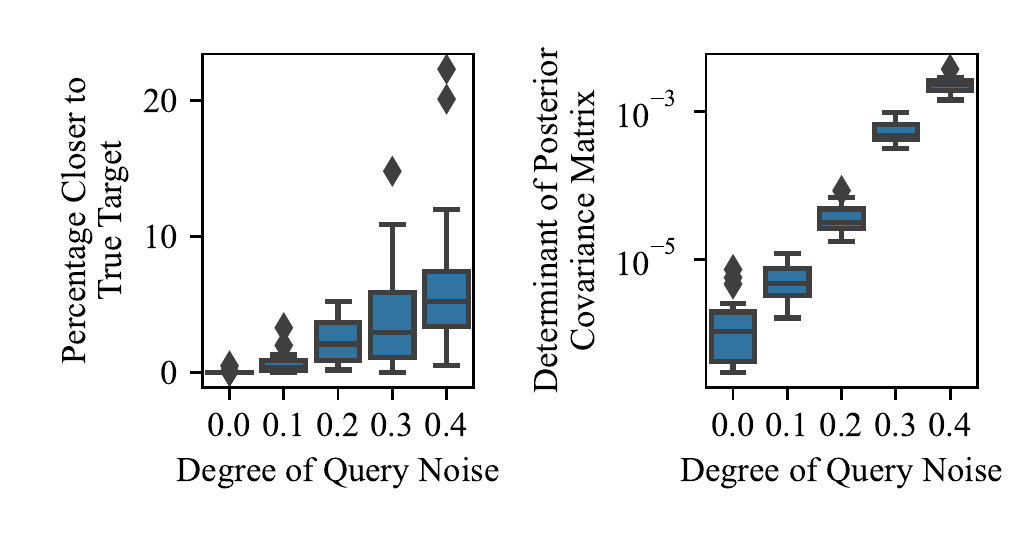}
    \caption{\textbf{Robustness of our method to query response noise.} Preference estimation under additive Gaussian noise. (Left) As noise is added, performance gradually degrades, while still remaining in the top estimation percentile. (Right) Uncertainty, as measured by the determinant of attribute posterior covariance, increases with increased noise. }
    \vspace{-0.1cm}
    \label{fig:noise_figure}
\end{figure}

\paragraph{Comparison to a Baseline Method}

We implement a baseline method from \cite{heim2019constrained}, which we call \textit{LSTM Baseline}. In this paper, the authors use an LSTM to estimate a latent vector that generates an image which satisfies a sequence of paired comparison constraints. In our approach, we separate preference estimation and mapping to the generative model latent space. For the sake of comparison, we extend the LSTM approach to estimate a low dimensional attribute vector that satisfies a given set of constraints. Further details in the implementation of this baseline are show in the Appendix \ref{experiment_details}. As shown in Figure \ref{fig:comparison_figure} and Table \ref{comparison_table}, our method with both random and active querying can estimate the target attributes with higher precision and efficiency. 

\begin{table*}[t!]
\begin{center}
\label{comparison_table}
\begin{tabular}{l c c c c c}

\hline
& MSE to G.T. Attribute $\downarrow$ & Percentage Closer ($\%$) $\downarrow$  & Constraint Satisfied ($\%$) $\uparrow$\\
\hline
LSTM & 5.86e-2 $\pm$ 5.60e-2 & 13.330 $\pm$ 10.390 & 83.83 $\pm$ 10.290 \\
Random  & 1.89e-1 $\pm$ 1.64e-2 & 5.731 $\pm$ 5.209 & 97.17 $\pm$ 2.420\\
Active (Best of N) & 5.37e-4 $\pm$ 6.41e-4 & 0.145 $\pm$ 0.139 & 98.50 $\pm$ 2.467\\
Active (Closed Form) & \textbf{1.69e-5} $\pm$ \textbf{2.35e-5} & \textbf{0.003} $\pm$ \textbf{0.001} & \textbf{99.83} $\pm$ \textbf{0.007} \\
\hline
\end{tabular}
\end{center}
\caption{ \textbf{Performance of Preference Estimation on Various Methods.} We quantitatively evaluate different preference estimation approaches after 30 queries on 20 experiment trials. \textit{MSE to Ground Truth} measures the Mean Squared Error of our final preference estimate in attribute space to the ground truth. \textit{Percentage Closer ($\%$)} shows the percentage of attribute vectors in attribute space that are closer to the ground truth than the preference estimate. \textit{Constraint Satisfied ($\%$)} shows the percentage of collected queries that are satisfied by our final preference estimation. For all metrics, our approaches (\textit{Random}, \textit{Active (Best of N)} and \textit{Active (Closed Form)}) outperform the \textit{LSTM} baseline, and \textit{Active (Closed Form)} outperforms all other methods across each metric. 
}
\end{table*}

\paragraph{Qualitative Performance of PrefGen}

Figure \ref{fig:preference_estimation_just_samples} shows how the image sampled for a given preference estimate changes as more queries are presented to the oracle. After each query we sample a latent vector $z_f = f(z_i, \mathbb{E}[a \mid Q_t])$, with attributes corresponding to the expected user preferences $\mathbb{E}[a \mid Q_t]$, and generate the corresponding image $G(z_f)$. Qualitatively, the attributes of the preference samples converge to the attributes of the target images. Figure \ref{fig:qualitative_gan_control_with_queries} shows two trial runs of our \textit{PrefGen} procedure. Qualitatively, it is clear that the preference estimate is converging on the target attributes using the information provided by each paired comparison query.

\paragraph{Robustness to Query Response Noise}

In most practical settings, various sources of noise can cause errors or inconsistencies in an oracle's responses to queries. Hence, it is desirable to build a system that is robust to such noise. To evaluate \textit{PrefGen}'s robustness to query noise, we purposely add Gaussian noise to our query likelihood as follows: $\sigma(k_q(||a - a_n||^2 - ||a - a_p||^2 + n))$ where $n \sim \mathcal{N}(0, \sigma_n)$ and $\sigma_n$ is the degree of noise we add. Figure \ref{fig:noise_figure} shows the percentage of attribute vectors from a large sample (1000) that are closer to the target attributes than the preference estimate after 30 queries for 20 trials. For each degree of noise, we tune $k_q$ using maximum likelihood estimation on a set of training triplets (see Appendix \ref{experiment_details} for more details). As the degree of noise increases, we note the performance of preference estimation degrades, but more importantly, it does not catastrophically fail. Additionally, we can quantify uncertainty in our posterior $p(a \mid Q_t)$ by computing the determinant of the covariance matrix of samples drawn from $p(a \mid Q_t)$, which shows that as we increase the amount of noise there is an increase in the uncertainty of our posterior.

\paragraph{Effectiveness of Closed Form Active Querying}

Section \ref{active_selection} discusses the importance of deploying a method which actively selects queries informative of an oracle's preference and effective for estimation with fewer responses. Figure \ref{fig:comparison_figure} clearly shows that actively selecting queries both drastically improves the efficiency of our preference estimation procedure and the final precision. Additionally, it is clear that the proposed \textit{Closed Form} active querying method outperforms the \textit{Best of N} approach. Figure \ref{fig:closed_form_comparison} shows that \textit{Closed Form} continues to outperform the \textit{Best of N} approach as one increases the number of candidate samples.

\subsection{PrefGen with Textual Attributes}

We demonstrate that it is possible to use \textit{PrefGen} to edit relative attributes described by text prompts. We find two directions in CLIP space, corresponding to the attributes ``anger'' and ``red hair''. For ``anger'' $t_n=\textit{``a person with a neutral expression"}$ and $t_p=\textit{``a person with an angry expression"}$ and for ``red hair" we use $t_n=\textit{``a person with dark hair"}$ and $t_p=\textit{``a person with red hair"}$.  Figure \ref{fig:intro_figure} shows \textit{PrefGen} in this 2D space formed by these attributes. We are not just interested in editing vectors within a GAN latent space, we also want to be able to edit real world images. To this end, we deploy GAN Inversion techniques, specifically use the Encoder4Editing \cite{encoder4editing}, to embed images into the latent space of our StyleGAN2 model, and edit those vectors.

\section{Discussion} 

In this work, we propose \textit{PrefGen}, a framework for controlling generative models using the information from pairwise comparisons. We demonstrate the success of this approach using StyleGAN2 on the task of human face editing. Our method works in conjunction with both a supervised attribute mapping and CLIP-based textual attribute mapping.  

A potential limitation of our experimental approach is that we deploy a simulated oracle in our experiments as a stand-in for a user's responses to queries instead of a user directly. However, we believe this is a reasonable assumption as we specifically construct our attribute embeddings to correspond to human interpretable features, and there is a substantial body of work studying the utility of pairwise comparisons for modeling user preferences in that setting \cite{david1963method, whittlesearch}. Another potential limitation is that we use linear directions in CLIP space to model relative attributes, however, there is no guarantee that these linear directions will encode the relative intensity of textual features. A potential line of future work is to better leverage joint text-image models like CLIP to model relative attributes.    

Despite their potential benefits, generative models also raise diverse ethical concerns that can be particularly acute when applied to images of people. Potential harmful uses include the creation of disinformation \cite{sedova2021ai} or ``deepfakes'' \cite{deep_fake_ref}. We urge developers applying this technology to carefully consider potential benefits and harms, and to pursue strategies to mitigate harms (e.g., \cite{mitigate_harms}) when appropriate.

\section{Acknowledgements}

We would like to thank the generous support and guidance of the members of the Georgia Tech SIPLab. This work was partially supported by Georgia Tech’s President’s Undergraduate Research Award (PURA), the NSF CAREER award CCF-1350954, and the NSF Research Experiences for Undergraduates (REU) program, and an ONR Grant. We would also like to thank Asa Harbin for helping with the manuscript.  

\clearpage

{\small
\bibliographystyle{ieee}
\bibliography{main_with_appendix}
}

\clearpage

\begin{appendices}

\section{Methods Details}

\section{Experiments Details \label{experiment_details}}

\subsection{Selecting an Optimal Noise Constant}

The performance of preference estimation is sensitive to certain hyperparameters. One such parameter is the noise constant $k_q$ which acts as a signal-to-noise ratio in our Bayesian approach to estimating user preferences. For some experiments we choose a constant $k_q = 10$ which was found to generally give strong performance. This is satisfactory when comparing different approaches where the degree of noise in user responses to queries remains fixed (e.g. the experiment showed in Figure \ref{fig:comparison_figure}). However, in Figure \ref{fig:noise_figure} we show results from an experiment investigating the impact of different degrees of noise on the preference estimation performance and uncertainty quantification capabilities of our methodology. For this experiment, it is necessary to tune a noise constant $k_q$ for each degree of noise. 

The way we tune $k_q$ for each degree of noise is to generate a small dataset ($\approx 1000$ examples) of attribute triplets $t = (a_a, a_p, a_n) \in T$ that satisfy $|a_a - a_p| + n < |a_a - a_n|$ where $n \sim N(0, \sigma_n)$. Here we are generating a dataset of triplets with varying degrees of noise, parameterized by the standard deviation $\sigma_n$. We want to select a noise constant $k_q$ for each degree of noise that maximizes the likelihood
\begin{equation}
    \sum_{(a_a, a_p, a_n) \in T} \sigma(k_q (||a_a - a_n||^2 - ||a_a - a_p||^2)).
\end{equation}
Intuitively, as the degree of noise in our dataset increases we want our model to become less confident in its estimate of the user's preferences. The way this is encoded is by decreasing the noise constant $k_q$ so that the $\sigma(k_q (||a_a - a_n||^2 - ||a_a - a_p||^2))$ goes closer to $0.5$ on average. This means that as we decrease $k_q$ we weight the impact of any given query lower, leading to slower convergence, and higher posterior volume. 

\subsection{Preference Estimation Experimental Procedure}

Given a generative model $G(z)$ (in our case a StyleGAN2 \cite{karras2019style} model) we simulate a user's responses to paired queries and attempt to estimate a user's preferences over a set of attributes $\mathcal{A}$. Let $T$ be the number of queries we present to a user. Let $a^* \in \mathcal{A}$ be the ground truth attributes that describe a user's preferences over a set of attributes $\mathcal{A}$. Our goal is to approximate $a^*$. 

For each trial we follow this procedure: 

\begin{enumerate}
    \item First we generate a set of target attributes $a^* \in \mathcal{A}$ sampled uniformly from $\mathcal{A}$. 
    \item We randomly sample an initial latent vector $z_i \in \mathcal{Z}$ from our StyleGAN2 latent space (actually $\mathcal{W}$ space of StyleGAN2, but we call it $\mathcal{Z}$ for convenience). 
    \item $Q_0 = \{ \}$
    \item For $T$ paired queries:
    \begin{enumerate}
        \item Sample a paired query $q = (a_a, a_b)$ of attributes using the desired sampling scheme (Active, Random, etc.).
        \item Generate the corresponding set of images for the query $q_x = (G(f(z_i, a_a)), G(f(z_i, a_b)))$
        \item Answer the query using a simulated oracle. 
        \item $Q_t = Q_{t-1} \cup q$
        \item Use the Stan modeling language \footnote{https://mc-stan.org/} to sample from the posterior $p(a \mid Q_t)$ using MCMC.
        \item Compute the current posterior estimate $\mathbb{E}[a \mid Q_t]$ of the oracle's preferences. 
        \item Compute the estimate latent vector $z_t = f(z_i, \mathbb{E}[a \mid Q_t])$. 
        \item Generate the estimate image $x_t = G(z_t)$. 
        \item Compute any desired evaluation metrics as a function of $z_t, x_t$ and $z_i$. 
    \end{enumerate}
\end{enumerate}

\subsection{LSTM Baseline Implementation Details} \label{lstm_baseline_information}

In the core paper, we implemented a variation of \cite{heim2019constrained}, which we call the LSTM Baseline. Here we describe the alterations that we made to their architecture for the sake of fair comparison.

\begin{figure*}[h]
    \centering
    \includegraphics[width=\columnwidth]{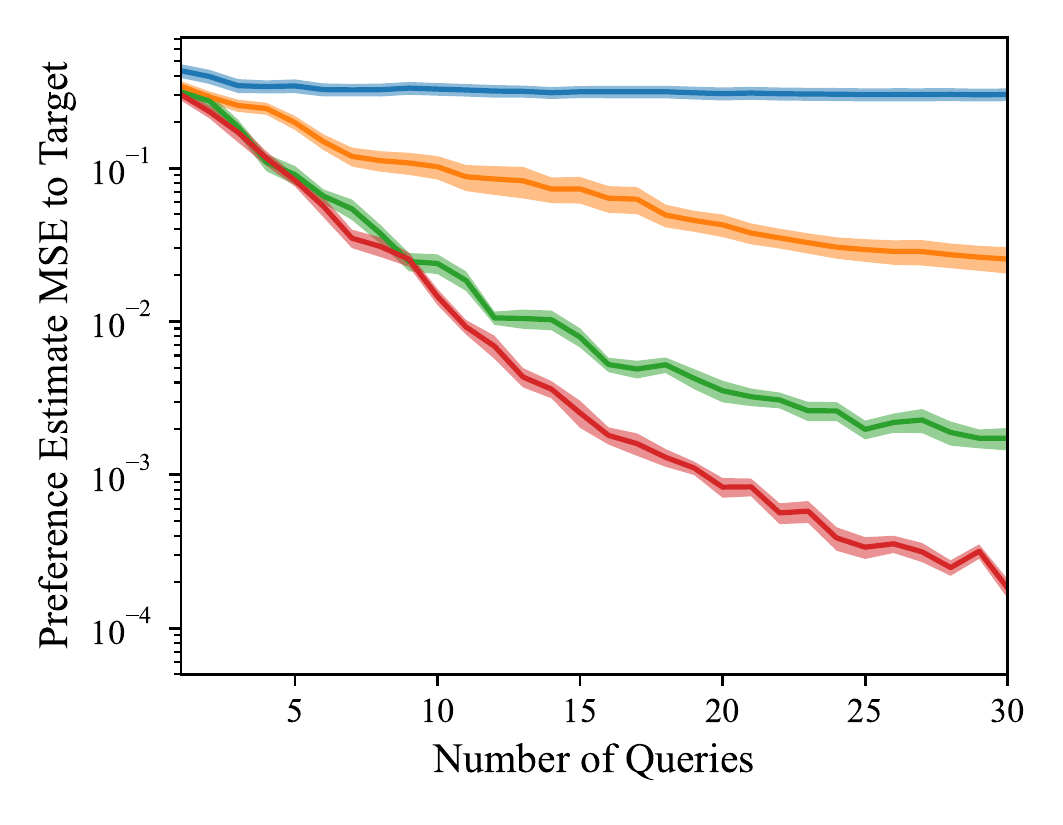}
    \includegraphics[width=\columnwidth]{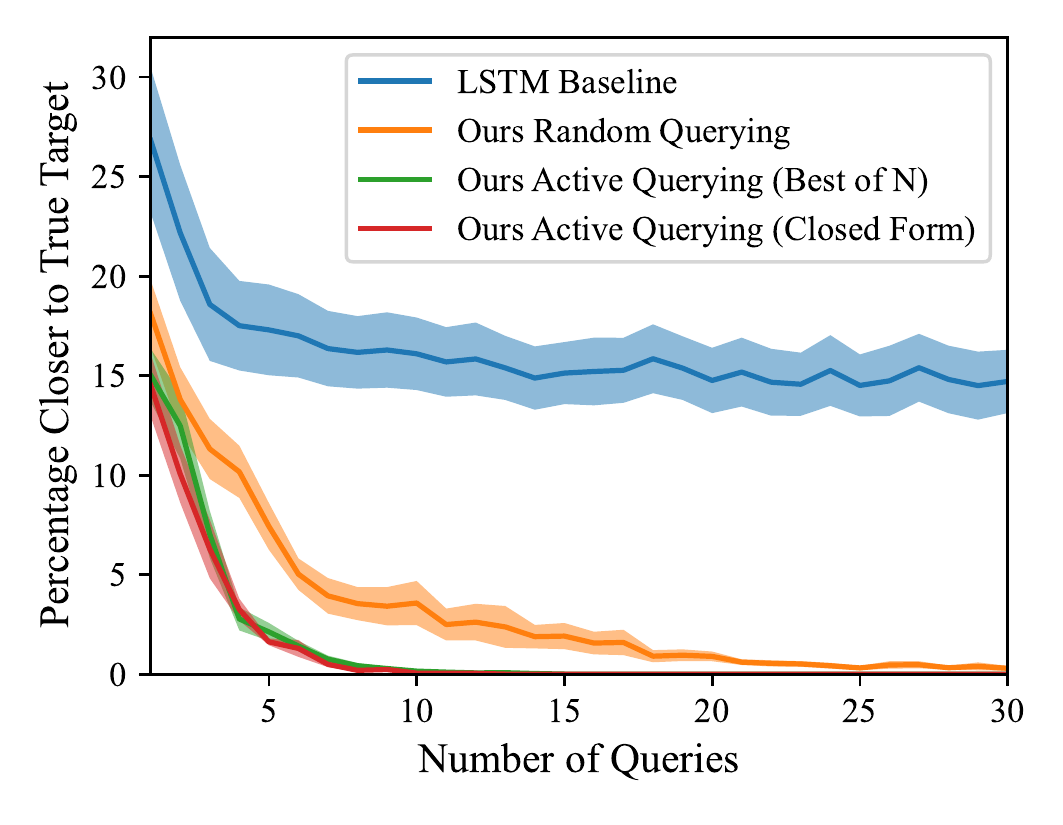}
    \caption{\textbf{Comparing the quantitative performance of various approaches to estimating user preferences from paired comparisons (4D Yaw, Pitch, Roll, and Age)}. This figure is similar to Figure \ref{fig:comparison_figure} from the core paper. We simulate an oracle's response to a sequence of queries for 20 trials, with 30 queries per trial. We search a 4D space formed by the age and the pose (yaw, pitch, and roll) of a face. (Left) MSE in attribute space of preference estimate after $t$ queries and the ground truth attribute $a^*$ for a given trial. (Right) Percentage of attribute vectors from a batch of 1000 that are closer to the target attributes than the estimate. According to both metrics, PrefGen is able to efficiently estimate preferences with high precision. As expected, active querying improves the efficiency of preference estimation as compared to randomly querying, and the \textit{Closed Form} approach, which leverages the continuous generative model latent space, significantly outperforms the \textit{Best of N} approach.}
    \label{fig:4d_quantitative_comparison}
\end{figure*}

\paragraph{The Original Approach}

In \cite{heim2019constrained}, the author wants to generate images from a GAN, given a set of pairwise constraints. This motivation is very similar to ours, however, their approach is substantially different. They use an LSTM architecture to estimate a latent vector $z_f$ that satisfies a given set of constraints given an input latent vector $z_i$. 

The LSTM architecture is defined as follows. The architecture has three core components. The first component is a \textit{read network} $\phi(x)$ that maps image constraints to a semantic space, this semantic space serves a similar function to our attribute space $\mathcal{A}$. The second component is a \textit{process network} that combines all of the constraints into a single representation, this serves a similar function to our Bayesian preference estimation procedure. The third component is a \textit{write network} which maps a final representation to an image, which is what our GAN generator does.

Let $\{c_1, \dots, c_n\}$ be a set of embedded constraints produced by the \textit{read network}. Each constraint $c_i = a_p \Vert a_n$ for a query $q_i = (a_p, a_n)$, where $\Vert$ corresponds to vector concatenation. The authors describe a \textit{process network} that takes in these constraints and an input latent vector $z$ and produce an output latent vector $z_f$ that can be fed into a generator to produce an image that satisfies constraints. The LSTM framework for the \textit{process network} is described below:

\begin{equation}
    h_t = LSTM(z, h_{t-1}^*)
\end{equation}
\begin{equation}
    e_{i, t} = c_i \cdot b_t
\end{equation}
\begin{equation}
    a_{i,t} = \frac{exp(e_{i,t})}{\sum_j^n exp(e_{j, t})}
\end{equation}
\begin{equation}
    r_t = \sum_i^n a_{i, t} c_i
\end{equation}
\begin{equation}
    h_t^* = h_t \Vert r_t
\end{equation}

As described in \cite{heim2019constrained} an input $z$ and the hidden state from the previous repetition are put through the LSTM. The next hidden state $h_t$ is then combined with each constraint $c_i$ using a dot product to produce a scalar value $e_{i, t}$ for each constraint. These values are then fed into a softmax function to obtain the scalars $a_{i, t}$ which is used in a weighted sum. The sum is the operation that combines the constraints. The final $r_t$ is concatenated to $h_t$ and is used as the input in the next processing iteration. This process is run for $p$ steps, to eventually produce $h_p^*$ which is fed into a fully connected layer to produce $s$ which is directly fed into the write network (generator). 

The authors train this framework end-to-end using a triplet loss and an adversarial training procedure.  

\paragraph{Our Experimental Changes}

In our paper, we do not need to train our system end-to-end. Instead, we segregate the procedure into two key components. First, we estimate user preferences over an attribute space using a Bayesian approach. Second, we outline a general procedure for modifying latent vectors so that they generate images that match those preferences. Our Bayesian preference estimation procedure serves a very similar function to the \textit{process network} in \cite{heim2019constrained}. The only key difference is that the process network in \cite{heim2019constrained} produces a vector $s$ which is fed directly into the \textit{write network} (generator). This requires the LSTM to both model a user's preferences given a set of constraints, and have an understanding of the structure of the generator latent space. Our preference estimation procedure is agnostic to any given latent representation and simply needs to find a vector in a low-dimensional (compared to a GAN latent space) attribute space that satisfies a set of constraints. For the sake of comparison, we modify the LSTM framework so that it simply needs to produce a low-dimensional attribute vector given a set of constraints. We can then plug this into our attribute mapping functions $f(z_i, a)$ to produce latent vectors that satisfy constraints. This allows us to directly compare the performance of our Bayesian preference estimation approach and the LSTM approach. 

Specific to this architecture, the only thing we change is to feed a zero vector as input into the LSTM instead of a latent vector $z$, while still giving the LSTM a set of constraints. We then treat the final output of the LSTM as an attribute vector $a$.

\begin{figure*}
    \centering
    \includegraphics[width=\columnwidth]{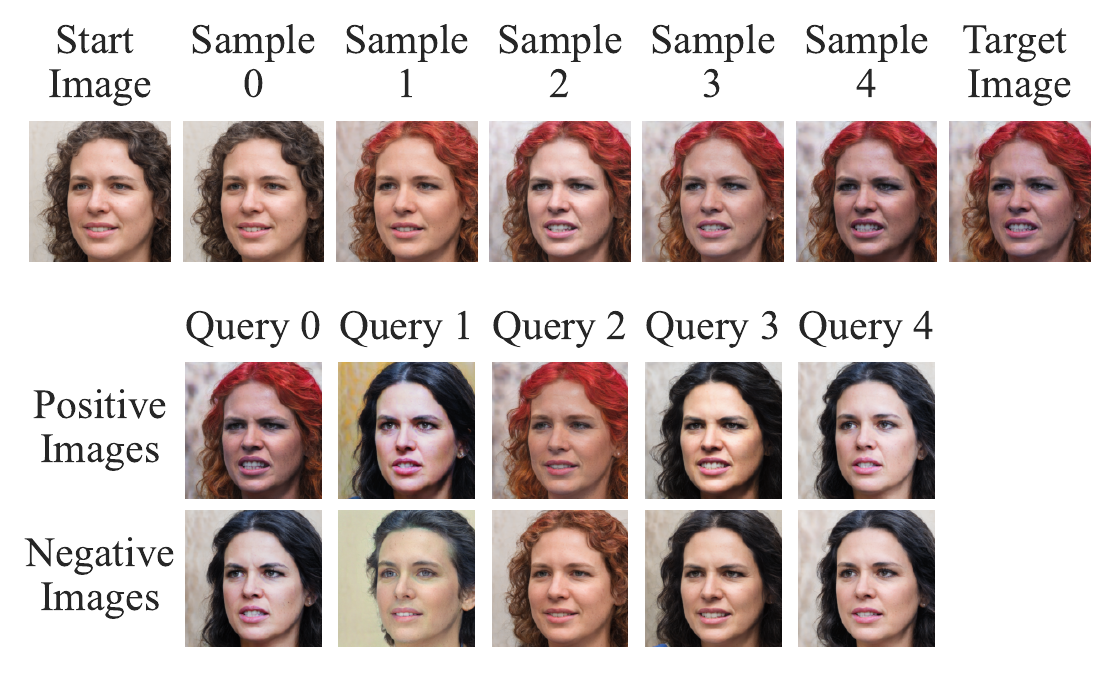}
    \hspace{0.2in}
    \includegraphics[width=\columnwidth]{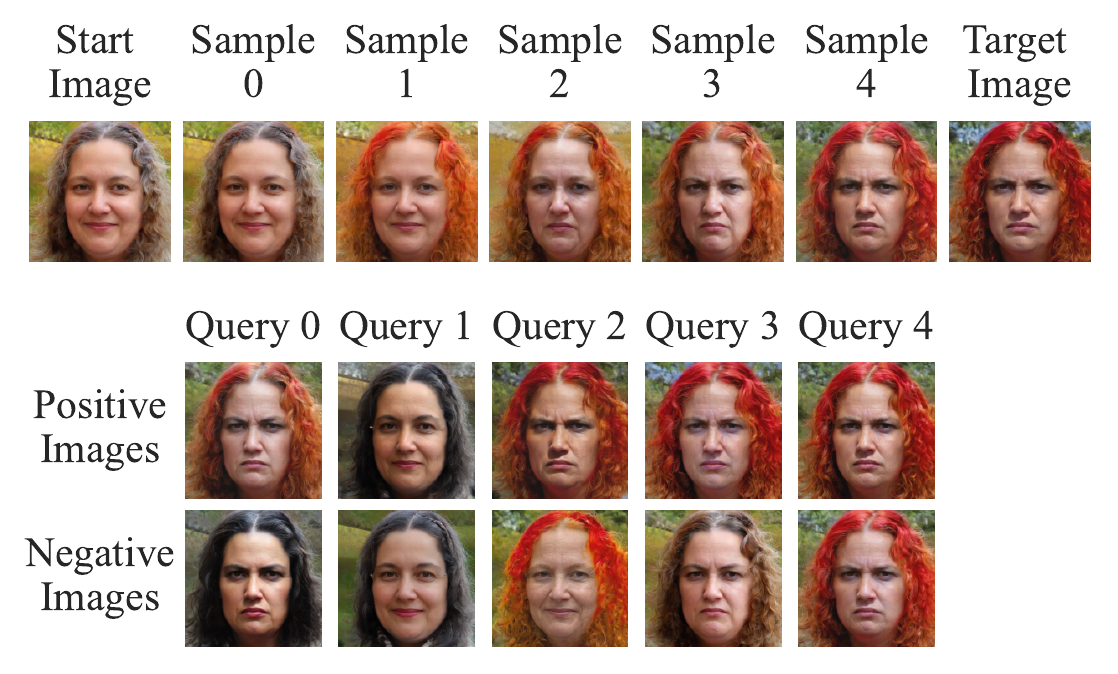}
    \caption{\textbf{PrefGen with CLIP.} Here we show two example trials of PrefGen in a 2D space formed by the textual attributes ``red hair" and ``anger".}
    \label{fig:clip_samples_and_queries}
\end{figure*}

\section{Further Results \label{further_results}}

\subsection{Supervised Mapping: Face Pose and Age Preference Estimation}

\paragraph{Qualitative: Face Pose and Age Preference Estimation}

\begin{figure*}
    \centering
    \includegraphics[width=\columnwidth]{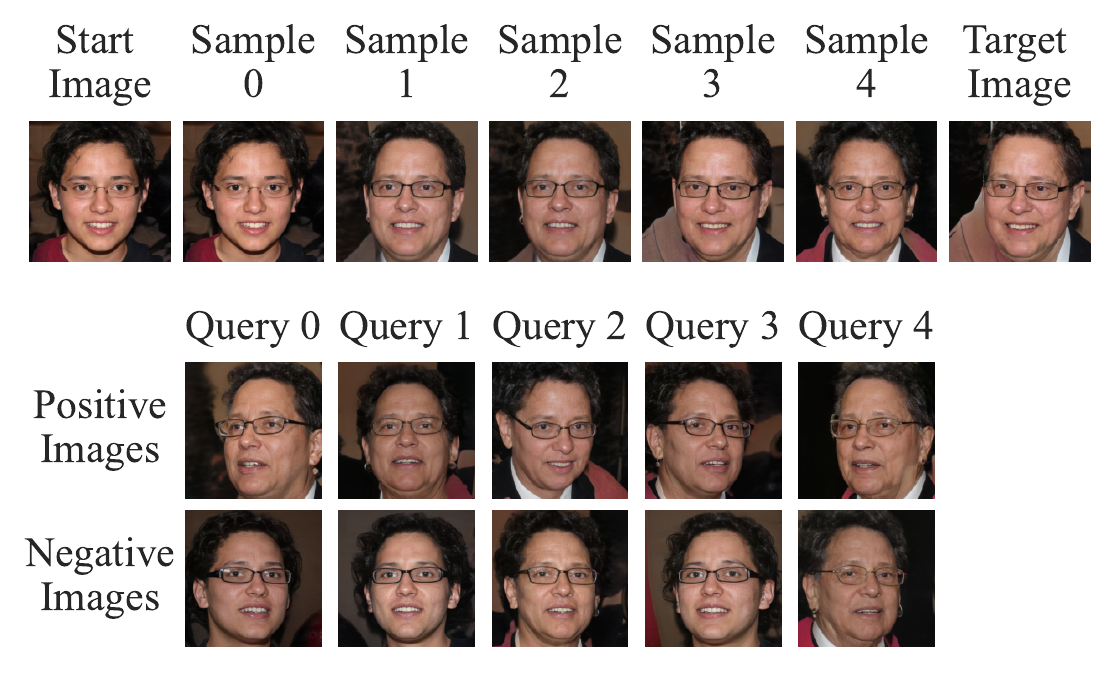}
    \hspace{0.2in}
    \includegraphics[width=\columnwidth]{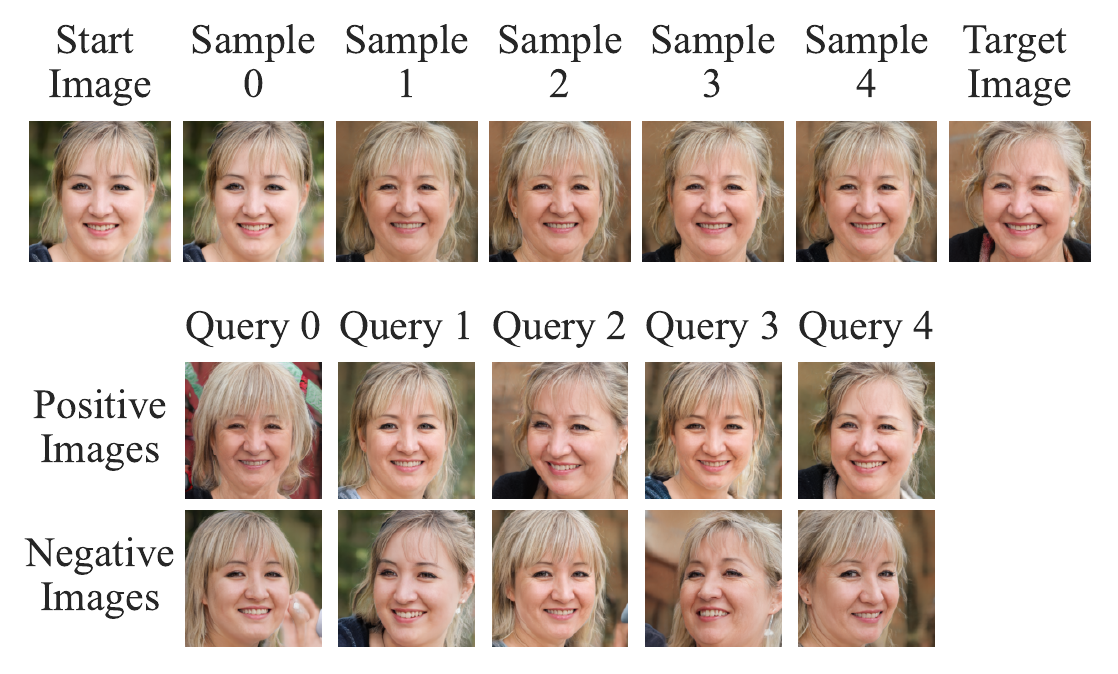}
    \newline
    \includegraphics[width=\columnwidth]{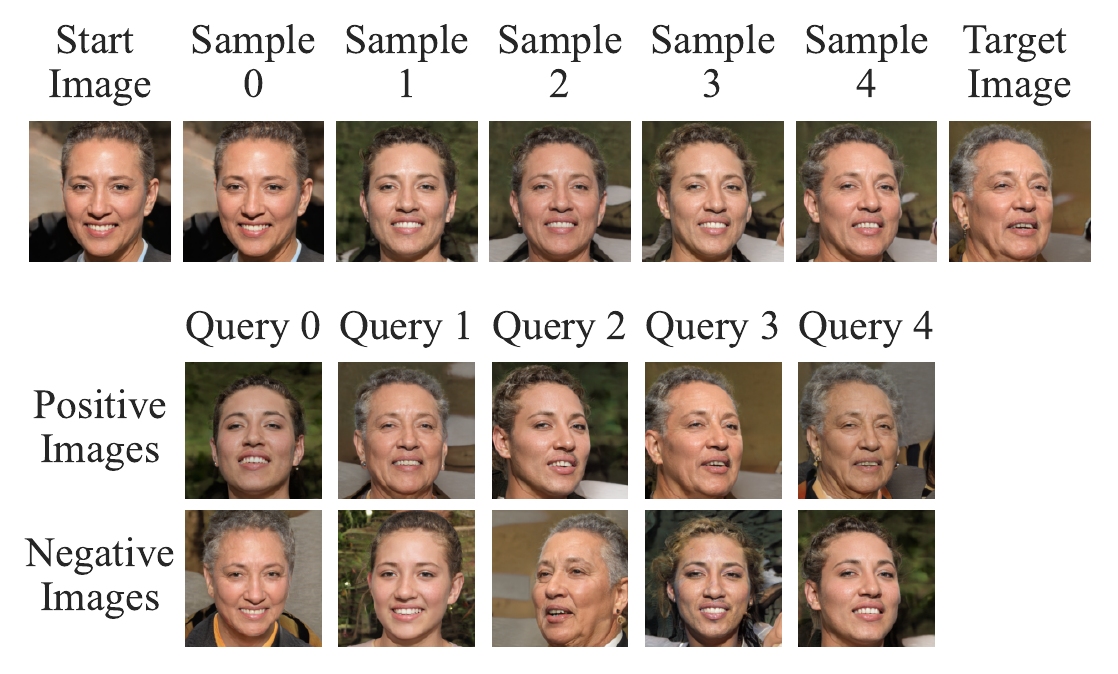}
    \hspace{0.2in}
    \includegraphics[width=\columnwidth]{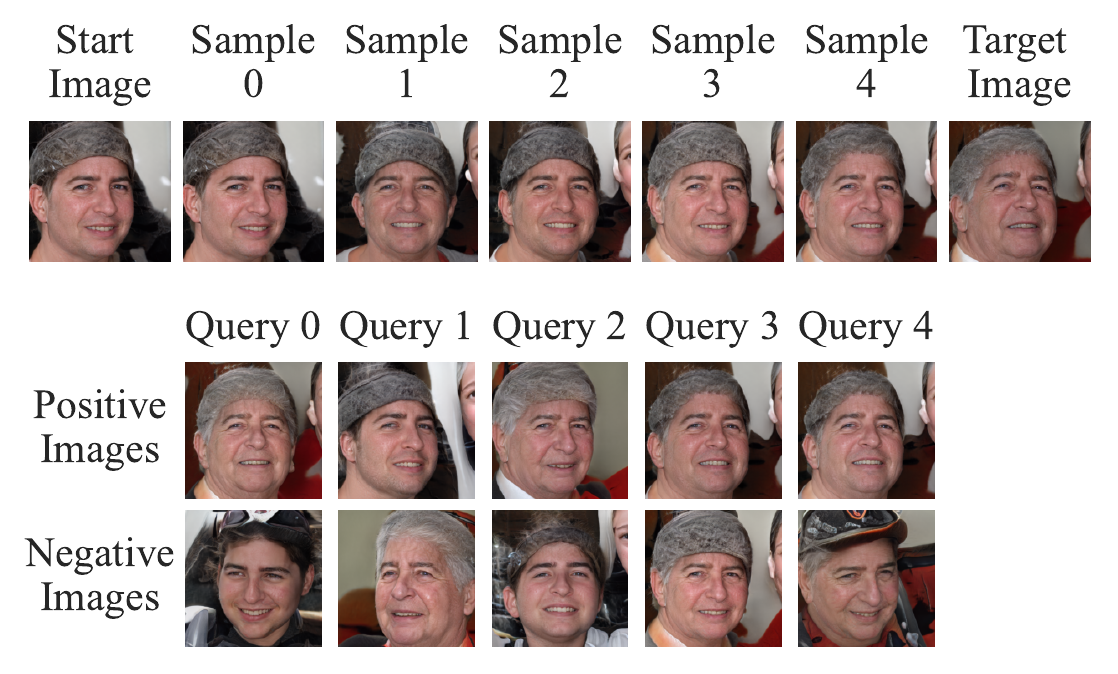}
    \newline
    \includegraphics[width=\columnwidth]{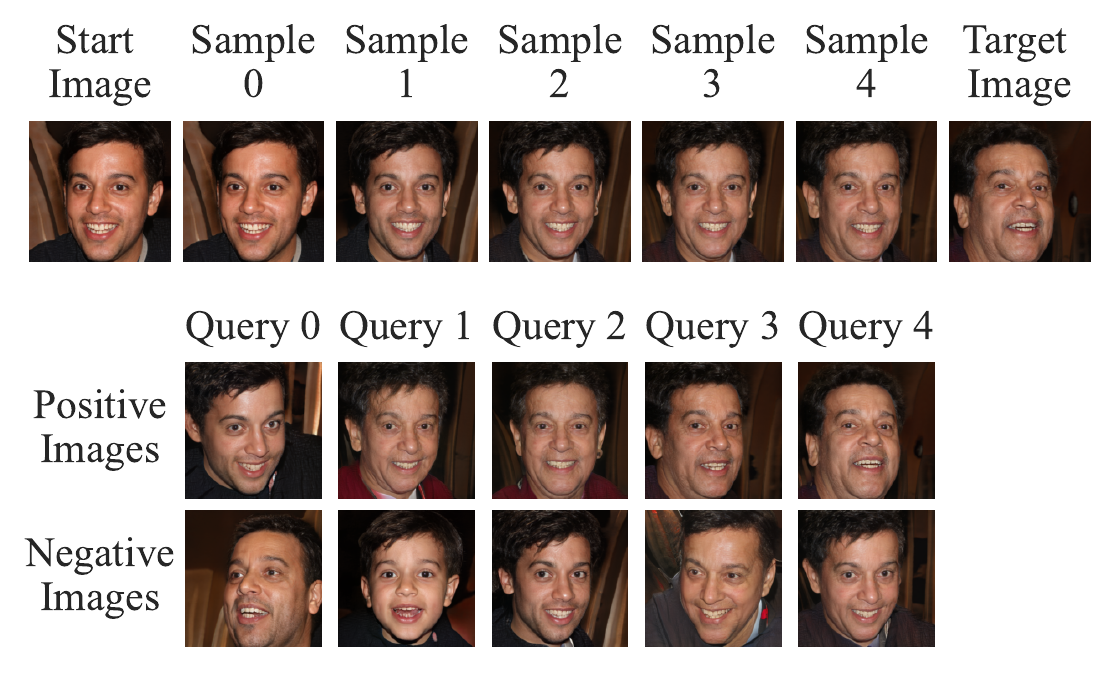}
    \hspace{0.2in}
    \includegraphics[width=\columnwidth]{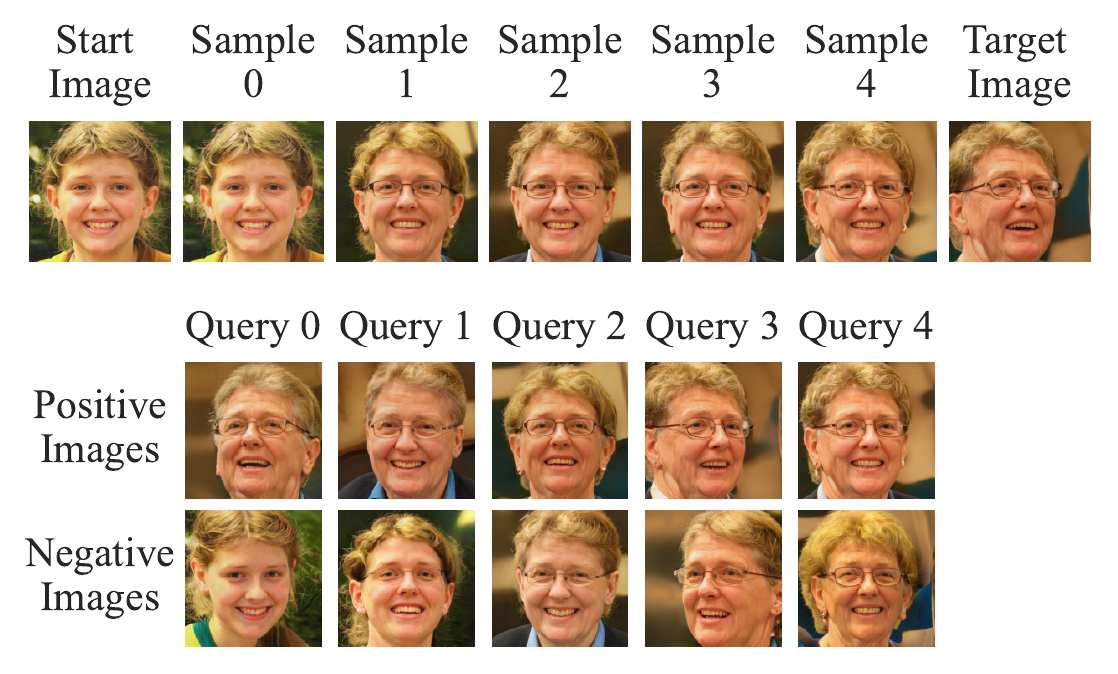}
    \newline
    \caption{\textbf{More Pose and Age Preference Estimation Examples}. Here are more examples of the preference estimation procedure similar to Figure \ref{fig:qualitative_gan_control_with_queries}. We demonstrate 6 trials of PrefGen in a 4D space formed by yaw, pitch, roll, and age of a human face. We show each query that is presented to the oracle, and the images generated by sampling from the posterior after each query. Sample images tend to qualitatively converge on the target. }
    \label{fig:full_page_localization_examples}
\end{figure*}

In Figure \ref{fig:qualitative_gan_control_with_queries} we showed qualitative examples of two trial runs of our preference estimation procedure. For each, we showed the queries that were presented to the oracle, the initial image, target image, and samples of how our estimate of the oracle's preferences changed over time. Because of a lack of space, we could not include more examples, but here we show several more. This is shown in Figure \ref{fig:full_page_localization_examples}. 

\paragraph{Quantitative: Face Pose and Age Preference Estimation}

In the core paper, we showed various quantitative results which demonstrated the effectiveness of our approach at estimating user preferences in a 2D attribute space formed by the left-right orientation of a face and the age of a face. However, it is important that our method generalizes to spaces with more attributes. Here we show parallel results in a 4D space formed by face age, yaw, pitch, and roll. These results are shown in Figure \ref{fig:4d_quantitative_comparison}. 

\subsection{CLIP Based Mapping}

\begin{figure}[H]
\vspace{-0.2in}
    \centering
    \includegraphics[width=3.25in]{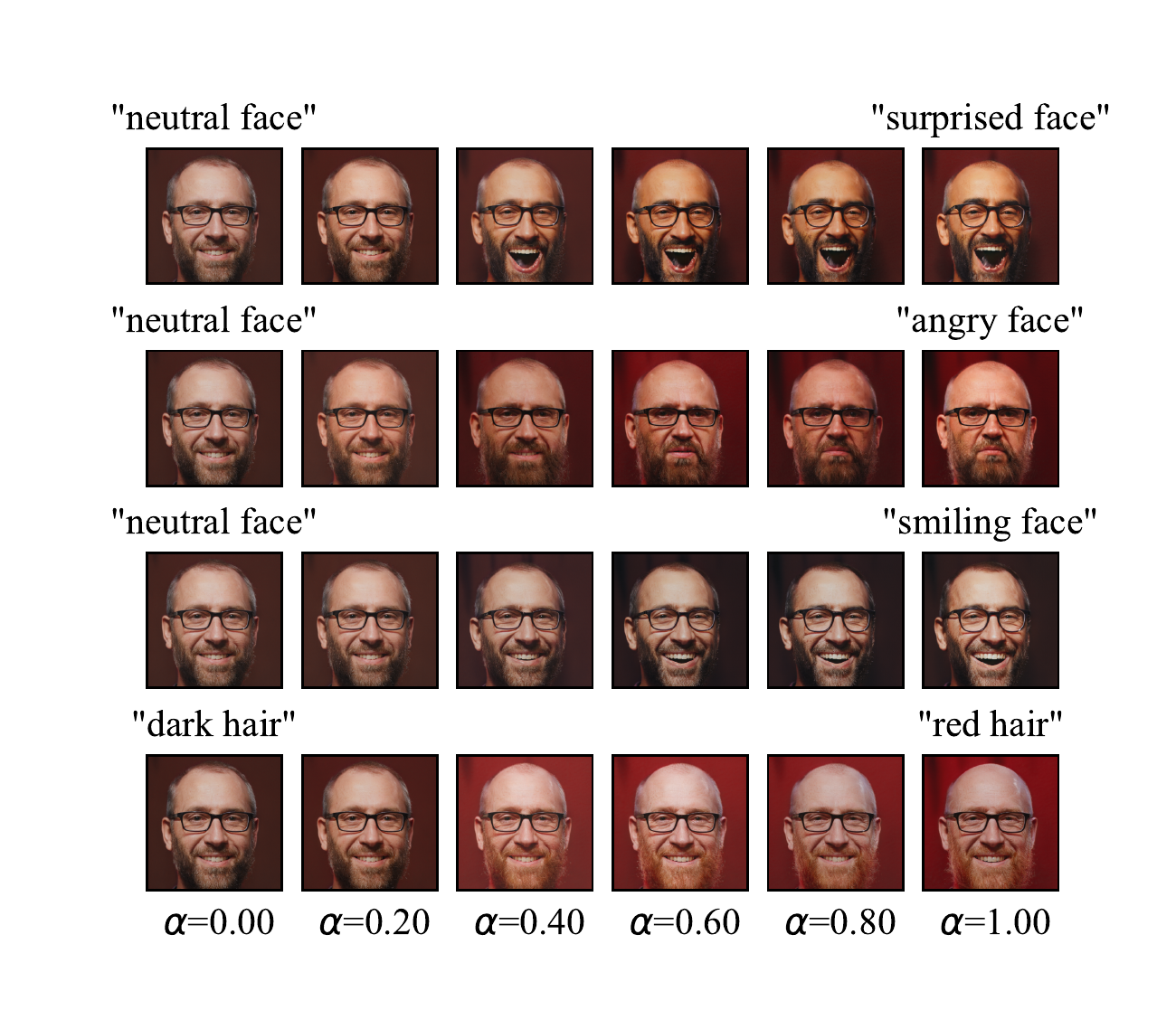}
    \caption{\textbf{Modeling Relative Attributes with CLIP.} It is possible to transform certain relative attributes of a face by inferring directions in CLIP space corresponding to those attributes. We can then use a gradient-based approach on the latent vector of an image until it produces images that embed to CLIP vectors with a certain intensity along this inferred relative attribute direction. }
    \label{fig:single_attributes}
\end{figure}

We used CLIP to infer directions that correspond to different relative textual attributes. In Figure \ref{fig:intro_figure} we showed a trial of PrefGen using an encoded image of Tom Cruise. In Figure \ref{fig:clip_samples_and_queries} we show two more example trials of the same features (i.e. ``red hair" and ``anger") with start images generated with random latent vectors. It is also possible to manipulate different types of relative textual attributes. Figure \ref{fig:single_attributes} shows the manipulation of several more relative attributes. 

\end{appendices}

\end{document}